\documentclass{article}

\usepackage{geometry}
\usepackage[utf8]{inputenc} 
\usepackage[T1]{fontenc}    
\usepackage{url}            
\usepackage{booktabs}       
\usepackage{amsfonts}       
\usepackage{nicefrac}       
\usepackage{microtype}      
\usepackage{float}
\usepackage{times}
\usepackage{epsfig}
\usepackage{graphicx}
\usepackage{amsmath}
\usepackage{amssymb}
\usepackage{algorithm}
\usepackage{algpseudocode}
\usepackage{xcolor,colortbl}
\usepackage{url}

\definecolor{Gray0}{rgb}{0.9,0.9,0.9}
\definecolor{Gray1}{rgb}{0.7,0.7,0.7}
\definecolor{Blue1}{rgb}{0.5,0.5,1.0}
\definecolor{Blue12}{rgb}{0.55,0.55,1.0}
\definecolor{Blue2}{rgb}{0.6,0.6,1.0}
\definecolor{Blue23}{rgb}{0.65,0.65,1.0}
\definecolor{Blue3}{rgb}{0.7,0.7,1.0}
\definecolor{Blue34}{rgb}{0.75,0.75,1.0}
\definecolor{Blue4}{rgb}{0.8,0.8,1.0}
\definecolor{Blue45}{rgb}{0.85,0.85,1.0}
\definecolor{Blue5}{rgb}{0.9,0.9,1.0}

\title{Dynamic Pose-Robust Facial Expression Recognition by Multi-View Pairwise Conditional Random Forests}

\author{Arnaud Dapogny$^1$\\
{\tt\small arnaud.dapogny@isir.upmc.fr}
\and
Kevin Bailly$^1$\\
{\tt\small kevin.bailly@isir.upmc.fr}
\and
S\'{e}verine Dubuisson$^1$\\
{\tt\small severine.dubuisson@isir.upmc.fr}\\\\
$^1$ Sorbonne Universit\'{e}s, UPMC Univ Paris 06, CNRS, ISIR UMR 7222\\ 4 place Jussieu 75005 Paris
}

\begin{document}

\date{}
\maketitle

\begin{abstract}
Automatic facial expression classification (FER) from videos is a critical problem for the development of intelligent human-computer interaction systems. Still, it is a challenging problem that involves capturing high-dimensional spatio-temporal patterns describing the variation of one's appearance over time. Such representation undergoes great variability of the facial morphology and environmental factors as well as head pose variations. In this paper, we use Conditional Random Forests to capture low-level expression transition patterns. More specifically, heterogeneous derivative features (e.g. feature point movements or texture variations) are evaluated upon pairs of images. When testing on a video frame, pairs are created between this current frame and previous ones and predictions for each previous frame are used to draw trees from Pairwise Conditional Random Forests (PCRF) whose pairwise outputs are averaged over time to produce robust estimates. Moreover, PCRF collections can also be conditioned on head pose estimation for multi-view dynamic FER. As such, our approach appears as a natural extension of Random Forests for learning spatio-temporal patterns, potentially from multiple viewpoints. Experiments on popular datasets show that our method leads to significant improvements over standard Random Forests as well as state-of-the-art approaches on several scenarios, including a novel multi-view video corpus generated from a publicly available database.
\end{abstract}

\section*{Introduction}

Over the last decades, automatic facial expression recognition (FER) has attracted an increasing attention \cite{zeng2009survey, sariyanidi2014automatic}, as it is a fundamental step of many applications such as human-computer interaction, or assistive healthcare technologies. The rationale behind those works is that decrypting facial expressions can serve as an unobtrusive way of analyzing one's underlying emotional state. Towards this end, a rich background literature has been developed by the psychological community in order to define models that can accurately and exhaustively represent facial expressions.

One of the most long-standing and widely used model is the discrete categorization proposed in the cross-cultural studies conducted by Ekman \cite{ekman1971constants}, which introduced six basic expressions that are universally recognized: \textit{happiness}, \textit{anger}, \textit{sadness}, \textit{fear}, \textit{disgust} and \textit{surprise}. This has been used to build as an underlying expression model for most attempts for \textit{prototypical} expression benchmarking and recognition scenarios \cite{lucey2010extended, yin20063d, yin2008high}, as the annotation process is quite intuitive. It can however show limitations for dealing with spontaneous expressions \cite{zhang2014bp4d}, as many of our daily affective behaviors are not covered by such prototypical emotions.

Another popular approach is the continuous dimensional representation of affect \cite{greenwald1989affective}, which consists in describing expressions in terms of a small number of latent variables rather than discrete categorical attributes. Perhaps one of the most widely used model is the valence/activation (relaxed \textit{vs.} aroused)/power (feeling of control)/expectancy (anticipation) model. This model is often further simplified as a $2D$ valence-activation representation. However, the projection of complex emotional states into such a low-dimensional embedding may result in loss of information. As a consequence, some expressions such as \textit{surprise} cannot be represented correctly whereas some others can not be separated efficiently (\textit{fear} \textit{vs.} \textit{anger}). Finally, the annotation process is less intuitive than with the categorical representation.

An alternative representation of facial expressions has been proposed under the form of the Facial Action Coding System (FACS) \cite{ekman1977facial}. Here, facial expressions are decomposed as a combination of $44$ facial muscle activations called Facial Action Units (AUs). AUs provide an intermediate face representation that is independent from interpretation and can in theory be combined in accordance with the so-called Emotional FACS (EMFACS) rules to describe any prototypical or spontaneous expression display. Unfortunately, the main drawback of this approach is that FACS-coding is generally cumbersome, and raters generally have to be highly trained, thus limiting the quantity of available data.

For those reasons, in this work we focus on categorical facial expression classification. However, there is nothing in our method that would prevent us from adapting our code to either the dimensional or FACS model, given appropriate data. More specifically, we aim at designing a FER system that:

\begin{itemize}
\item is able to reliably distinguish subtle expressions (e.g. \textit{anger} or \textit{sadness}). Because using dynamics of the expression helps disentangle the factors of variation \cite{cohn2006foundations}, such system needs to exploit the temporal variations in videos rather than trying to perform recognition on still images. For that matter, we focus on combining the benefits of frame-based and dynamic classifications;
\item is robust to contextual factors (e.g. lighting conditions) and can perform recognition from arbitrary viewpoints, depending on head pose variation or camera position;
\item can be learned from available data and work in real-time on a standard computer with any basic webcam plugged in. Particularly, we do not use high-resolution $3D$ face scans because many approaches \cite{vieriufacial, ben20144} working on such data seem to perform poorly when applied on low-resolution, noisy consumer sensors such as the Kinect. In addition, depth information may be unavailable in many applicative scenarios.
\end{itemize}

\section{Related work}\label{related}

In this section we review recent works addressing FER from video. On the one's hand, recent approaches for FER from a frontal view can be divided in frame-based systems and dynamic ones. On the other hand, multi-view FER is generally performed statically.

\subsection{Frame-based FER}\label{frameFER}

The first category of FER systems are the so-called frame-based classifiers.
For instance, Khan \textit{et al.} \cite{khan2012human} propose a human vision-inspired framework that applies classification upon Pyramid Histogram of Orientation Gradients (PHOG) features from salient facial regions. Happy \textit{et al.} \cite{happyautomatic} extract prominent facial patches from the position of facial landmarks. A subset of discriminative salient patches can then be used for FER.

This category of approaches typically aims at outputting an expression prediction for each separate frame. Hence, they can generally be applied to classify each frame of a video without pre-segmentation. Unfortunately, they also suffer from a number of drawbacks. First, they typically require frame-level annotations for training, which can be a time-consuming process. Secondly, frame-level approaches essentially ignore a part of the information as they do not exploit the temporal evolution of the features. They also do not use the temporal correlations at the semantic level (e.g. is it plausible to predict \textit{sadness} immediately after having recognised \textit{happiness}?). Recently, Meguid \textit{et al.} \cite{abd2014fully} obtained promising results by accumulating hybrid RF/SVM predictions into histograms computed using a sliding window.

In order to disentangle facial morphology from expression, other approaches explicitly normalize each image  w.r.t. a neutral face representation. Mohammadi \textit{et al.} \cite{mohammadi2014non} use a constrained smoothed $l_0$-norm sparse decomposition to infer facial expressions from differences of face images. However, the neutral face has to be provided beforehand, limiting the applicability of these methods. In order to circumvent this issue, the so-called dynamic FER methods typically make use of spatio-temporal information.

\subsection{Dynamic FER}\label{videoFER}

Dynamic information of facial expressions can be used in several ways: (a) at the feature-level, by using spatio-temporal image descriptors, and/or (b) at the semantic level, by modelling relationships between expressions or between successive phases (\textit{onset}, \textit{apex} and \textit{offset}) of facial events. Generally speaking, effectively extracting suitable representations from spatio-temporal video patterns is a challenging problem as expressions may occur with various offsets and at different paces. There is no consensus either on how to combine those representations flexibly enough so as to generalize on unseen data and possibly unseen temporal variations. Common approaches employ spatio-temporal descriptors defined on fixed-size windows, optionally at multiple resolutions. Examples of such features include the so-called LBP-TOP \cite{zhao2007dynamic,hayat2012evaluation} and HOG3D \cite{klaser2008spatio} descriptors, which are spatio-temporal extensions of LBP and HOG features respectively. Authors in \cite{shojaeilangari2014multi} use histograms of local phase and orientations. However, those kind of representations may lack the capacity to generalize to facial events that differ from training data on the temporal axis.

Approaches trying to address (b) aim at establishing relationships between high-level features and a sequence of latent states. Wang \textit{et al.} \cite{wang2013capturing} integrate temporal interval algebra into a Bayesian network to capture complex relationships among facial muscles. Such approaches generally require explicit dimensionality reduction techniques such as PCA or $k$-means clustering for training. In addition, training at the sequence level reduces the quantity of available training and testing data as compared to frame-based approaches, as there is only one expression label per video.

In a previous work \cite{dap2015trans}, we trained Random Forests upon pairs of images representing expression transition patterns. Those forests were conditioned on the expression label of the first frame to help reducing the variability. Although we obtained promising results for dynamic FER from frontal views of the videos, the proposed approach did not handle head pose variations.

\subsection{Multi-view FER}\label{mvFER}

Many approaches for multi-view FER consist in training a single classifier to describe every viewpoint. Zheng \textit{et al.} \cite{zheng2010emotion} introduce a regional covariance matrix representation of face images to infer static facial expressions on a corpus constructed from the BU-3DFE database \cite{yin20063d} with $35$ different head poses up to $\pm 45$ yaw and $\pm 30$ pitch. Tariq \textit{et al.} \cite{tariq2012multi} address the same problem by using a translation invariant sparse coding of dense SIFT features. Eleftheriadis \textit{et al.} \cite{eleftheriadis2015discriminative} employ discriminative shared Gaussian processes to implicitly exploit the redundancy between multiple views of the same expressive images. However, such approach can struggle to capture the variability of the facial expressions when the number of training samples becomes important.

Alternatively, it is possible to learn a projection of a non-frontal views of a face image on a frontal one. Recently, Vieriu \textit{et al.} \cite{vieriufacial} proposed to project $3D$ data of the face onto a head pose-invariant $2D$ representation. The visible fraction of the projected face is then used within a voting scheme to decipher the expression. FER can thus be performed using an off-the-shelf algorithm. In addition, the authors were able to perform FER under a broader range of poses, up to $\pm 90$ yaw and $\pm 60$ pitch. However, the proposed method requires high-resolution $3D$ face data that may not necessarily be available in multiple human-computer interaction scenarios, for instance when using images acquired with commercial lower-resolution depth sensors.

Last but not least, some other works choose to learn one specific classifier per face view. During testing, the head pose is first estimated, then the best pose-specific expression classifier is applied. For instance, Moore \textit{et al.} learn multi-class SVMs upon LBP features for multiple viewpoints. Such approaches offer several advantages over the previous ones: first, learning classifiers upon separate and more homogeneous face view data allows to considerably reduce the variability. As a consequence the classifiers can, in theory, more efficiently capture the subtle facial deformations between the expressions. Secondly, the runtime is the same as in the case of a single frontal view classifier, which may be a critical point for systems that try to project a given view on a frontal one. Finally, splitting the training data offers the advantage to reduce the memory usage, which can be important for learning on large databases. Those methods also face some impediments, such as the fact that (a) they require a reliable facial landmark alignment and head pose estimation, and (b) it implies dividing the data into several subsets. Nevertheless, (a) is barely a problem given that recent advances \cite{xiong2013supervised,ren2014face,xiong2015global} for face alignment provide excellent results for head poses up to $\pm 45$ yaw and $\pm 30$ pitch, which is sufficient for most human-computer applications. Furthermore, (b) can be circumvented by the use of $3D$ face scans \cite{yin2008high} from which we can generate a large corpus of videos for training multi-view dynamic classifiers.

\section{Overview of proposed approach}\label{overview}

\begin{figure*}[htb]
\centering
\includegraphics[width=\linewidth]{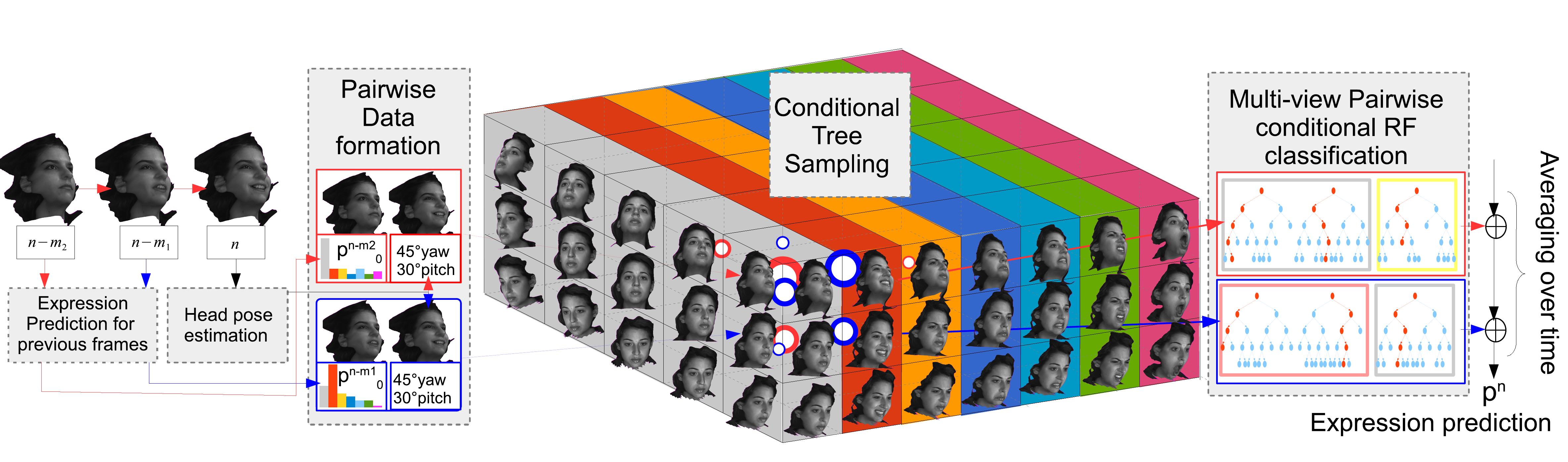}
\caption{Flowchart of our MVPCRF FER method. When evaluating a video frame indexed by $n$, pairs are created between this current frame and previous frames $n-m_1,n-m_2,...$. Randomized trees trained upon a pairwise dataset are then drawn conditionally to head pose estimation as well as expression probabilities for the previous frames. Finally, predictions outputted for each pair are averaged over time to give rise to an expression probability $p^n$ for the current frame. This prediction is used as a tree sampling distribution for classifying the following frames. Best viewed in color.}
\label{oobaccweps}
\end{figure*}

In this paper, we introduce the Multi-View Pairwise Conditional Random Forest (MVPCRF) algorithm, which is a new formulation for training trees using low-level heterogeneous static (spatial) and dynamic (spatio-temporal derivative) features within the Random Forest (RF) framework. Conditional Random Forests have recently been used by Dantone \textit{et al.} \cite{dantone2012real} as well as Sun \textit{et al.} \cite{sun2012conditional} in the field of facial alignment and human pose estimation, respectively. They generated collections of trees for specific, quantized values of a global variable (such as head pose \cite{dantone2012real} and body torso orientation \cite{sun2012conditional}) and used prediction on this global variable to draw dedicated trees, resulting in more accurate predictions. As depicted on Figure \ref{oobaccweps}, we propose to condition pairwise trees on specific expression labels to reduce the variability of ongoing expression transitions from the first frame of the pair to the other one. Furthermore, similarly to \cite{dantone2012real}, we can further condition the pairwise trees on a head pose estimation to add robustness towards head pose variations. When evaluating a video frame, each previous frame of the sequence is associated with this current frame to give rise to a pair. Pairwise trees are thus drawn from the dedicated PCRFs w.r.t. prediction for the previous frame. \textit{In Extenso}, a head pose estimate can be used to draw trees from MVPCRFs for pose-robust FER. Finally, predictions outputted for each pair are averaged over time to produce a robust prediction for the current frame. Contributions of this work are listed below.

\begin{itemize}
\item A method for training pairwise random trees upon high-dimensional heterogeneous static and spatio-temporal derivative feature templates, with a conditional formulation that reduces the variability of the transition patterns.
\item An extension of the traditional RF model averaging, that consists in averaging over time pairwise predictions to flexibly handle temporal variations.
\item A method for performing multi-view dynamic FER that consists in further conditioning pairwise trees on a pose estimate. A tree sampling probability distribution is constructed from the data to allow a continuous shift between the pose-specific PCRF models.
\item A new multi-view video corpus, that includes a method for aligning facial feature points on non-frontal sequences using an off-the-shelf feature point tracker. We provide source code for generating the data from $3D$ models using an available database \cite{yin2008high}.
\item A complete PCRF framework that performs fully automatic dynamic FER with a multi-view extension, that can work on low-power engines thanks to an efficient implementation using integral feature channels.
\end{itemize}

The rest of the paper is organized as follows: in Section \ref{staticrf} we describe our adaptation of the RF framework to learn expression patterns on still images from high-dimensional, heterogeneous (geometric/appearance) features. In Section \ref{technical} we present the MVPCRF framework for capturing spatio-temporal patterns that represent facial expressions from multiple viewpoints. In Section \ref{datapreparation} we explain how we generate a multi-view dynamic database for training and testing the models. In Section \ref{Experiments} we show how our PCRF algorithm improves the accuracy on several FER datasets compared to a static approach as well as state-of-the-art approaches. In Section \ref{expmonov} we report results from frontal view FER and in Section \ref{expMVPCRF} we report accuracy for non frontal head poses, showing that our MVPCRF formulation substantially increases the robustness to pose variations. In Section \ref{RealTime} we report the ability of our framework to run in real-time. Finally, we give concluding remarks on our MVPCRF for FER and discuss upcoming perspectives.

\section{Random Forests for FER}\label{staticrf}

\subsection{Random Forests}\label{staticpred}

Random Forests (RFs) is a popular learning framework introduced in the seminal work of Breiman \cite{breiman2001random}. They have been used to a significant extent in computer vision and for FER tasks in particular due to their ability to handle high-dimensional data such as images or videos as well as being naturally suited for multiclass classification tasks. They combine random subspace and bagging methods to provide performances similar to the most popular machine learning methods, such as SVM or neural networks.

RFs are classically built from the combination of $T$ decision trees grown from bootstraps sampled from the training dataset. In our implementation, we downsample the majority classes within the bootstraps in order to enforce class balance. As compared to other methods for balancing RF classifiers (\textit{i.e.} class weighting and upsampling of the minority classes), downsampling leads to similar results while substantially reducing the computational cost, as training is performed on smaller data subsets.

Individual trees are grown using a greedy procedure that involves, for each node, the measure of an impurity criterion $H_{\phi,\theta}$ (which is traditionally either defined as the Shannon entropy or the Gini impurity measurement) relatively to a partition of the images $x$ with label $l \in \mathcal{L}$, that is induced by candidate binary split functions $\{\phi,\theta\} \in \Phi$. More specifically, we use multiple parametric feature templates to generate multiple heterogeneous split functions, that are associated with a number of thresholds $\theta$. In what follows, by abuse of notations we will refer to $\phi^{(i)}$ as the $i^{th}$ feature template and $k^{(i)}$ as the number of candidates generated from this template. The ``best" binary feature among all features from the different templates (\textit{i.e.} the one that minimizes the impurity criterion $H_{\phi,\theta}$) is set to produce a data split for the current node. Then, those steps are recursively applied for the left and right subtrees with accordingly rooted data until the label distribution is homogeneous, where a leaf node is set. This procedure for growing trees is summarized in Algorithm \ref{treeGrowing}.


\begin{algorithm}
\caption{Tree Growing algorithm \texttt{treeGrowing}}
\label{treeGrowing}
\textbf{input:} images $x$ with labels $l$, root node $n$, number of candidate features $\{k^{(i)}\}_{i=1,2,3}$ for templates $\{\phi^{(i)}\}_{i=1,2,3}$
\begin{algorithmic}
\If{image labels are homogeneous with value $l_0$}
\State set node as terminal, with probabilities $p_t$ to $1$ for $l_0$, 0 elsewhere
\Else
\State generate an empty set of split candidates $\Phi$
\ForAll{feature templates $i$}, 
\State generate a set $\Phi^{(i)}$ of $k^{(i)}$ candidates $\{\phi^{(i)},\theta\}$
\State $\Phi \leftarrow \Phi \cup \Phi^{(i)}$
\EndFor
\For{$\{\phi,\theta\} \in \Phi$}
\State compute the impurity criterion $H_{\phi,\theta}(x)$
\EndFor
\State split data w.r.t. $\arg\min_{\{\phi,\theta\}}\{H_{\phi,\theta}(x)\}$ in left and right subsets $x_l$ and $x_r$
\State create left ($n_l$) and right ($n_r$) children of node $n$
\State call \texttt{treeGrowing}($x_l$,$n_l$,$\{k^{(i)}\}_{i=1,2,3}$)
\State call \texttt{treeGrowing}($x_r$,$n_r$,$\{k^{(i)}\}_{i=1,2,3}$)
\EndIf
\end{algorithmic}
\end{algorithm}

During evaluation, an image $x$ is successively rooted left or right of a specific tree $t$ according to the outputs of the binary tests, until it reaches a leaf node. The tree thus returns a probability $p_t(l | x)$ which is set to either $1$ for the represented class, or to $0$. Prediction probabilities are then averaged among the $T$ trees of the forest (Equation \eqref{probasRF}).

\begin{equation}\label{probasRF}
   p(l | x) = \frac{1}{T}\sum \limits_{t=1}^{T}{p_t(l | x)}
\end{equation}

Note that the robustness of the RF prediction framework comes from (a) the strength of individual trees and (b) the decorrelation between those trees. By growing trees from different bootstraps of available data and with the random subspace algorithm (e.g. examining only a subset of features for splitting each node) we generate weaker, but less correlated trees that provide better combination predictions than CART or C4.5 procedures \cite{quinlan2014c4}.

\subsection{Heterogeneous feature templates}\label{staticfeat}

Feature templates $\phi^{(i)}$ include both geometric (\textit{i.e.} computed from previously aligned facial feature points) and appearance features. Each of these templates have different input parameters that are randomly generated during training by uniform sampling over their respective variation range. Also, during training, features are generated along with a set of candidate thresholds $\theta$ to produce binary split candidates. For each template $\phi^{(i)}$, the upper and lower bounds are estimated from the training data and candidate thresholds are drawn from uniform distributions within this range.

We use two different geometric feature templates which are generated from the set of facial feature points $f(x)$ aligned on image $x$ with SDM \cite{xiong2013supervised}. The first geometric feature template $\phi^{(1)}_{a,b}$ is the distance between feature points $f_a$ and $f_b$, normalized w.r.t. inter-ocular distance $iod(f)$ for scale invariance (Equation \ref{pointdist}).

\begin{equation}\label{pointdist}
   \phi^{(1)}_{a,b}(x) = \frac{||f_a - f_b||_2}{iod(f)}
\end{equation}

Because the feature point orientation is discarded in feature $\phi^{(1)}$ we use the angles between feature points $f_a$, $f_b$ and $f_c$ as our second geometric feature $\phi^{(2)}_{a,b,c,\lambda}$. In order to ensure continuity for angles around $0$, we use the cosine and sine instead of the raw angle. Thus, $\phi^{(2)}$ outputs either the cosine or sine of angle $\widehat{f_af_bf_c}$, depending on the value of the boolean parameter $\lambda$ (Equation \eqref{pointangle}):

\begin{equation}\label{pointangle}
   \phi^{(2)}_{a,b,c,\lambda}(x) = \lambda\cos(\widehat{f_af_bf_c}) + (1-\lambda)\sin(\widehat{f_af_bf_c})
\end{equation}

We use Histogram of Oriented Gradients (HOG) as our appearance features for their descriptive power and robustness to global illumination changes. In order to ensure fast feature extraction, we use integral feature channels as introduced in \cite{dollar2009integral}. First, images are rescaled to a constant size of $250 \times 250$ pixels. Then, we compute horizontal and vertical gradients on the image and use these to generate $9$ feature maps, the first one containing the gradient magnitude, and the $8$ remaining correspond to a 8-bin quantization of the gradient orientation. Then, integral images are computed from these feature maps to output the $9$ feature channels. Thus, we define the appearance feature template $\phi^{(3)}_{\tau,ch,s,\alpha,\beta,\gamma}$ as an integral histogram computed over channel $ch$ within a window of size $s$ normalized w.r.t. the inter-ocular distance. Such histogram is evaluated at a point defined by its barycentric coordinates $\alpha$, $\beta$ and $\gamma$ w.r.t. vertices of a triangle $\tau$ defined over feature points $f(x)$. Also, we store the gradient magnitude in the first channel to normalize the histograms. Thus, HOG features can be computed with only $4$ access to the integral channels (plus normalization).

However, the proposed static RF does not use the dynamics of the expressions, which is the purpose of the next section.

\section{Learning temporal patterns from multiple viewpoints}\label{technical}

\subsection{Learning PCRF with heterogeneous derivative feature templates}\label{PCRFtrain}

\begin{figure*}[ht]
\centering
\includegraphics[width=\linewidth]{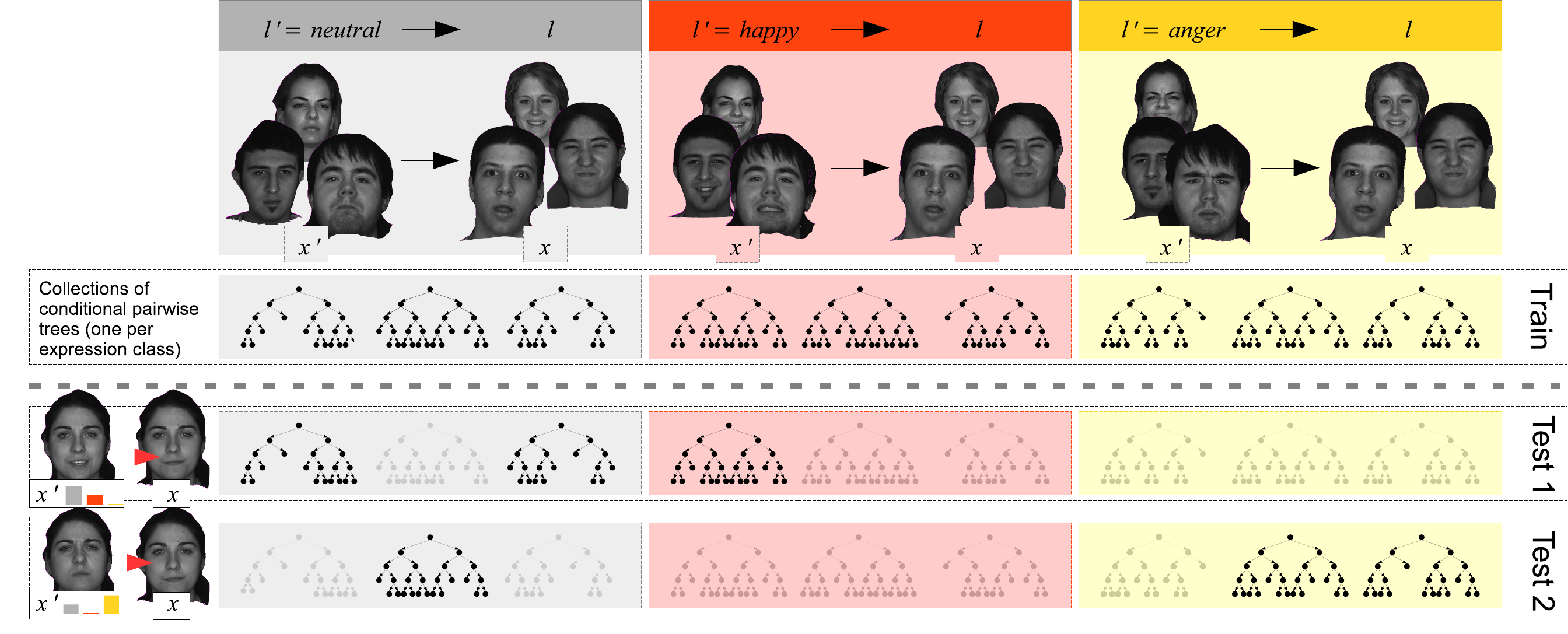}
\caption{Expression recognition from pairs of images using PCRF. Expression probability predictions of previous images are used to sample trees from dedicated pairwise tree collections (one per expression class) that are trained using subsets of the (pairwise) training dataset, with only examples of ongoing transitions from a specific expression towards all classes. The resulting forest thus outputs an expression probability for a specific pair of images.}
\label{maot}
\end{figure*}

In this section we now consider pairs of images ($x'$,$x$) to train trees $t$ that aim at outputting probabilities $p_t(l|x',x,l')$ of observing label $l(x)=l$ given image $x'$ and subject to $l(x')=l'$, as shown in Figure \ref{maot}. More specifically, for each tree $t$ among the $T$ trees of a RF dedicated to transitions starting from expression label $l'$, we randomly draw a fraction of subjects ${\tilde{\mathcal{S}} \subset \mathcal{S}}$. Then, for each subject $s \in \tilde{\mathcal{S}}$ we randomly draw images $x'_s$ that specifically have label $l'$. We also draw images $x_s$ of every label $l$ and create the pairs ($x'_s$, $x_s$) with label $l$. Note that the two images of a pair need to belong to the same subject, but not necessarily to the same video. Indeed, we create pairs from images sampled across different sequences for each subject to cover all sorts of ongoing transitions. We then balance the pairwise bootstrap by downsampling the majority class w.r.t. the pairwise labels. Eventually, we construct tree $t$ by calling Algorithm \ref{treeGrowing}. Those steps are summarized in Algorithm \ref{rfprocedure}.

\begin{algorithm}
\caption{Training a PCRF}
\label{rfprocedure}
\textbf{input:} images $x$ with labels $l$, number of candidate features $\{k^{(i)}\}_{i=1,...,6}$ for templates $\{\phi^{(i)}\}_{i=1,...,6}$
\begin{algorithmic}
\ForAll{$l' \in \mathcal{L}$}
\For{$t=1$ to $T$}
\State randomly draw a fraction $\tilde{\mathcal{S}} \subset \mathcal{S}$ of subjects
\State $pairs \leftarrow \{\}$
\ForAll{$s \in \tilde{\mathcal{S}}$}
\State draw samples $x'_s$ with label $l'$
\State draw samples $x_s$ for each label $l$
\State create pairwise data ($x'_s$, $x_s$) with label $l$
\State add element ($x'_s$, $x_s$) to $pairs$
\EndFor
\State balance bootstrap $pairs$ with downsampling
\State create new root node $n$
\State call \texttt{treeGrowing}($pairs$,$n$,$\{k^{(i)}\}_{i=1,...,6}$)
\EndFor
\EndFor
\end{algorithmic}
\end{algorithm}

\begin{figure}[ht]
\centering
\includegraphics[width=0.7\linewidth]{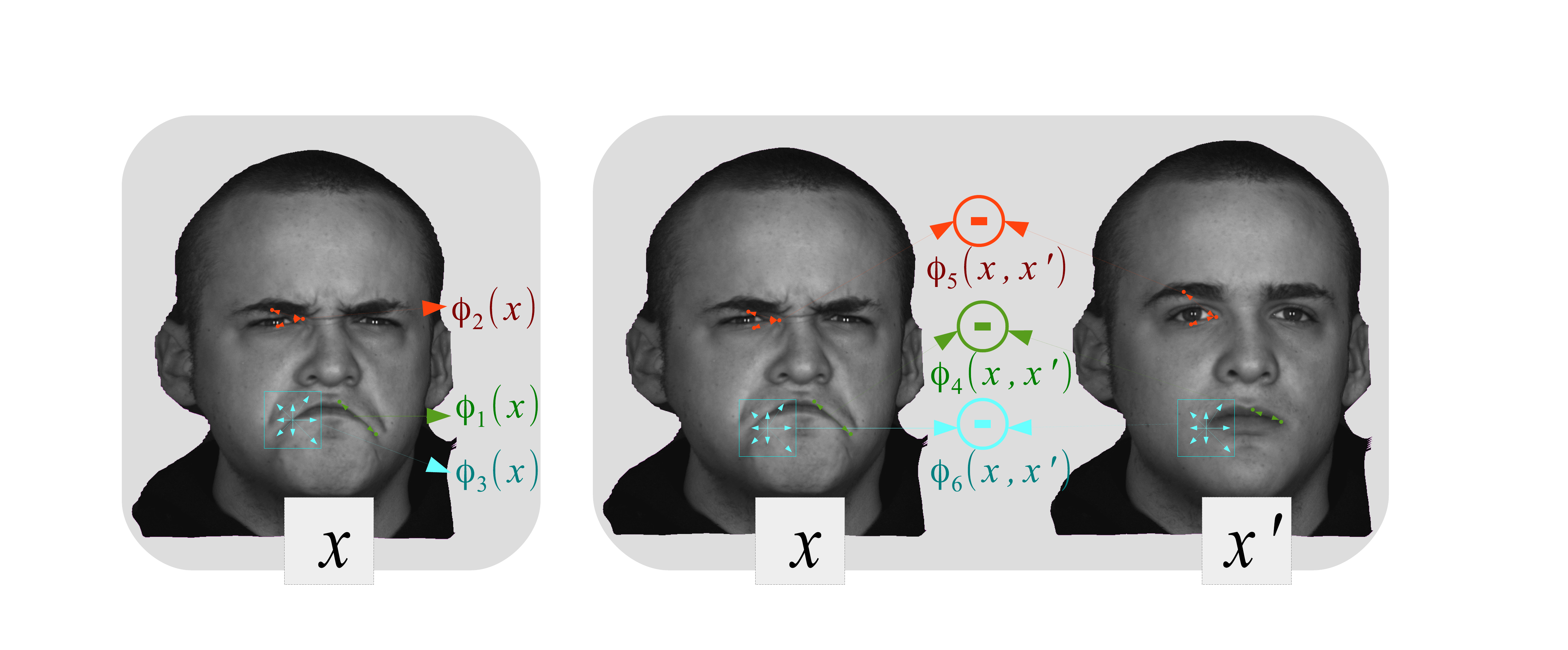}
\caption{Static (left) and pairwise (right) feature templates.}
\label{figures}
\end{figure}

As shown on Figure \ref{figures}, candidates for splitting the nodes are generated from an extended set of $6$ feature templates $\{\phi^{(i)}\}_{i=1,...,6}$, three of which being the static features described in Section \ref{staticrf}, that are applied to the second image $x$ of the pair ($x'$, $x$), for which we want to predict facial expressions. The three remaining feature templates are dynamic features defined as the derivatives of static templates $\phi^{(1)}$, $\phi^{(2)}$, $\phi^{(3)}$ with the exact same parameters. Namely, we have:

\begin{equation}\label{schema4}
\begin{cases}
\phi^{(1)}_{a,b}(x',x) & = \phi^{(1)}_{a,b}(x) \\
\phi^{(2)}_{a,b,c,\lambda}(x',x) & = \phi^{(2)}_{a,b,c,\lambda}(x) \\
\phi^{(3)}_{\tau,ch,s,\alpha,\beta,\gamma}(x',x) & = \phi^{(3)}_{\tau,ch,s,\alpha,\beta,\gamma}(x) \\
\phi^{(4)}_{a,b}(x',x) & = \phi^{(1)}_{a,b}(x) - \phi^{(1)}_{a,b}(x')\\
\phi^{(5)}_{a,b,c,\lambda}(x',x) & = \phi^{(2)}_{a,b,c,\lambda}(x) - \phi^{(2)}_{a,b,c,\lambda}(x') \\
\phi^{(6)}_{\tau,ch,s,\alpha,\beta,\gamma}(x',x) & = \phi^{(3)}_{\tau,ch,s,\alpha,\beta,\gamma}(x) - \phi^{(3)}_{\tau,ch,s,\alpha,\beta,\gamma}(x') \\

\end{cases}
\end{equation}

As in Section \ref{staticrf}, thresholds for the derivative features $\phi^{(4)}$, $\phi^{(5)}$, $\phi^{(6)}$ are drawn from uniform distributions with new dynamic template-specific ranges estimated from the pairwise dataset.

Note that, as compared to a static RF, a PCRF model is extended with new derivative features that are estimated from a pair of images. When applied on a video, predictions for several pairs are averaged over time in order to produce robust estimates of the probability predictions.

\subsection{Model averaging over time}\label{MAOT}

We denote by $p^n(l)$ the prediction probability of label $l$ for a video frame $x^n$ . For a purely static RF classifier this probability is given by Equation \eqref{pgeneric}:

\begin{equation}\label{pgeneric}
   p^n(l) = \frac{1}{T}\sum \limits_{t=1}^{T}{p_t(l | x^n)}
\end{equation}

In order to use spatio-temporal information, we apply pairwise RF models to pairs of images ($x^m$, $x^n$) with $\{x^m\}_{m=n-1,...,n-N}$ the previous frames in the video. Those pairwise predictions are averaged over time to provide a new probability estimate $p^n$ that takes into account past observations up to frame $n$. Thus, if we do not have prior information for those frames the probability $p^n$ becomes:

\begin{equation}\label{pfull}
   p^n(l) = \frac{1}{NT}\sum\limits_{m=n-N}^{n-1} \sum\limits_{t=1}^{T} p_t(l|x^m,x^n)
\end{equation}

In what follows, Equation \eqref{pgeneric} and Equation \eqref{pfull} will respectively be referred to as the \textit{static} and \textit{full models}. Trees from the full model are likely to be stronger that those of the static one since they are grown upon an extended set of features. Likewise, the correlation between the individual trees is also lower thanks to the new features as well as the averaging over time. However, spatio-temporal information can theoretically not add much to the accuracy if the variability of the pairwise data points is too large.

In order to decrease this variability, we assume that there exists a probability distribution $p_0^m(l')$ to observe the expression label $l'$ at frame $m$. Note that those probabilities can be set to purely static estimates (which is necessarily the case for the first video frames) or dynamic predictions estimated from previous frames. A comparison between those approaches can be found in Section \ref{expmonov}. In such a case, for frame $m$, pairwise trees are drawn from the trees collections (each one being conditioned to one expression label for the first frame of the pair) by sampling the distribution $p_0^m$, as shown in Figure \ref{maot}. More specifically, for each expression label $l'$ we randomly select $\mathcal{N}(l')$ trees over a PCRF model dedicated to transitions that start from expression label $l'$, trained with the procedure described in Section \ref{PCRFtrain}. We denote $p_t^{l'}$ the probabilities outputted by a pairwise tree $t$ conditioned on label $l'$. Equation \eqref{pfull} thus becomes:


\begin{equation}\label{pcond}
   p^n(l) = \frac{1}{NT}\sum\limits_{m=n-N}^{n-1} \sum\limits_{l' \in \mathcal{L}} \sum\limits_{t=1}^{\mathcal{N}(l')} p_t(l|x^m,x^n,l')
\end{equation}

Where $\mathcal{N}(l') \approx Tp_0^m(l')$ and $T = \sum_{l' \in \mathcal{L}}\mathcal{N}(l')$ are the number of trees dedicated to the classification of each transition, which can be set in accordance with CPU availability. In our experiments, we will refer to Equation \eqref{pcond} as the \textit{conditional model}. This conditional formulation helps to reduce the variability of the derivative features for each specialized pairwise RF. When predicting expression for a frame of a video, we can effectively use robust sequence-level expression estimates by averaging over time predictions conditioned on multiple, independent previous frames. Section \ref{Experiments} shows that using PCRF models for FER leads to significant improvements over both static and full models.

\subsection{Multi-view formulation}\label{MVPCRF}

In order to design a pose-robust recognition framework, we propose to condition the  models w.r.t a head pose estimate $\omega(x^n)$ for frame $n$. For that matter we quantize the pose space $\Omega$ in $k = \Gamma \times B$ pose bins $\{\Omega_i = \Omega_{\gamma_i,\beta_i}\}_{i = 1,...,k}$, that are defined around yaw and pitch angles $\gamma_i$ and $\beta_i$, respectively. We can thus rewrite Equation \ref{pgeneric} as a static multi-view model (MVRF):

\begin{equation}\label{mvpcrfprobas}
   p^n(l) = \frac{1}{T}\sum\limits_{\Omega_i \in \Omega} \sum\limits_{t=1}^{\mathcal{N}(\Omega_i)} p_t(l|x^n,\Omega_i)
\end{equation}

At frame $n$, the head pose $\omega(x^n)$ is estimated first using an off-the-shelf posit algorithm \cite{dementhon1995model}. Then, for each pose bin $\Omega_i$, a number $\mathcal{N}(\Omega_i)$ of trees are selected based on a pose sampling probability distribution $\mathcal{P}_{\Omega_i}(\omega^n)$ that we construct from the training data repartition, as it will be explained in Section \ref{datapreparation}. Furthermore, we adapt Equation \eqref{pcond} by conditioning the expression-conditional model on pose estimation $\omega(x^n)$ (Equation \eqref{mvpcrfprobas}):

\begin{equation}\label{mvpcrfprobas}
   p^n(l) = \frac{1}{T}\sum\limits_{m=n-N}^{n-1} \sum\limits_{\Omega_i \in \Omega} \sum\limits_{l' \in \mathcal{L}} \sum\limits_{t=1}^{\mathcal{N}(l',\Omega_i)} p_t(l|x^n,x^m,\Omega_i,l')
\end{equation}

In what follows, we refer to this model as the \textit{multi-view PCRF} (MVPCRF) model. In this formulation, for computing the pairwise probability between frames $n$ and $m$, we first estimate the head pose for frame $n$. Then, for each pose bin $\Omega_i$ and expression label $l'$, we select a number of trees equal to $\mathcal{N}(l',\Omega_i)$ (Equation \eqref{mvpcrfdensity1}):

\begin{equation}\label{mvpcrfdensity1}
\mathcal{N}(l',\Omega_i) \approx T\mathcal{P}_{\Omega_i}(\omega(x^n))p^m_0(l')
\end{equation}

Where $p^m_0(l')$ is the probability of expression label $l'$ for frame $m$. The number of trees allocated to classify each transition is thus:

\begin{equation}\label{mcrftreenum}
T = \sum\limits_{\Omega_i \in \Omega} \sum\limits_{l' \in \mathcal{L}}\mathcal{N}(l',\Omega_i)
\end{equation}

Note that the tree sampling distribution proposed in Equation \ref{mvpcrfdensity1} supposes that the head pose estimate do not vary that much between frames $n-N$ and $n$. Should that be the case, MVPCRF can be trained from pairs of images from different viewpoints. It also assumes the independence of head pose and expression prior, which is not problematic for training on posed expression data. However, such assumption may not hold for spontaneous datasets for which expressions as \textit{surprise} or \textit{fear} may involve specific head motion (e.g. recoil). In such case, prior conditionals may be estimated from the training corpus beforehand. Also, as stated in \cite{dantone2012real,sun2012conditional} using conditional models usually involves one major pitfall, which lies in the reduction of the number of training examples used to train each separate classifier. This is barely a problem for the training of a PCRF model, as naturally many examples of each ongoing transition can be sampled from the datasets. Furthermore, for the MVPCRF model we can generate a new database that contains a large number of training examples for each pose bin using the high-resolution $3D$-models from the BU-4DFE database \cite{yin2008high}.

\section{Multi-view database generation}\label{datapreparation}

\begin{figure*}[htbf]
\centering
\begin{minipage}{1.0\textwidth}
\includegraphics[width=\linewidth]{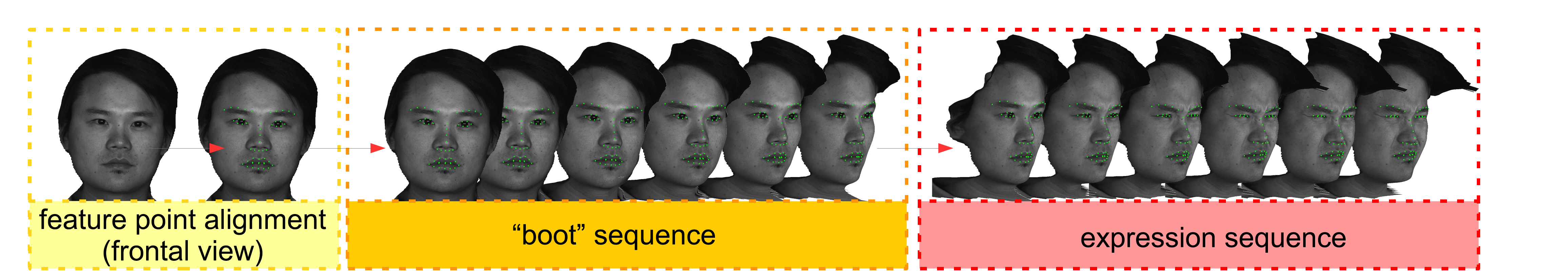}
\caption{Boot process for multi-view data generation with aligned feature points}
\label{boot}
\end{minipage}\hfill

\centering
\begin{minipage}{0.5\textwidth}
\centering
\includegraphics[width=\linewidth]{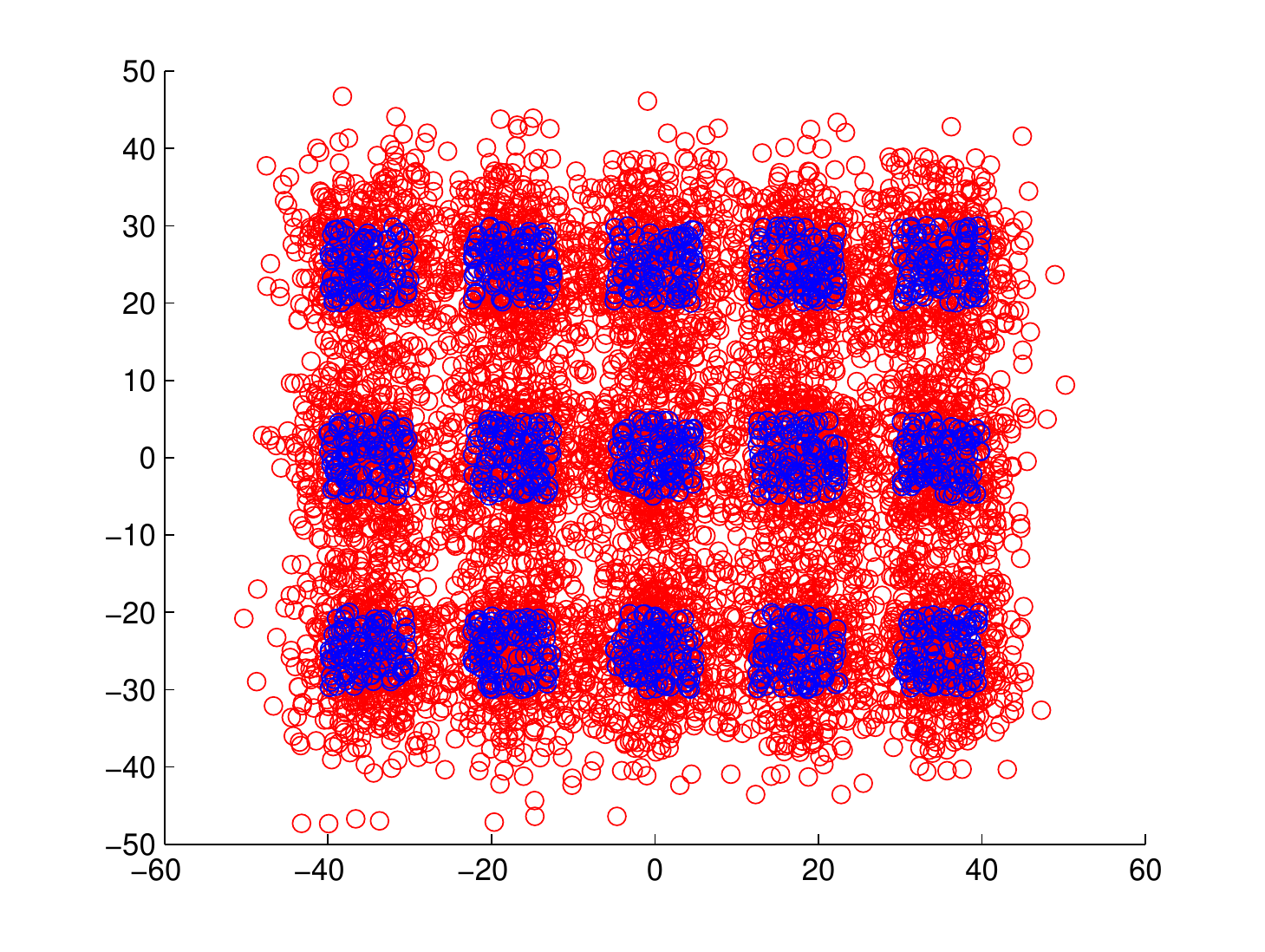}
\caption{Data repartition across the $15$ generated pose bins. Blue circles: angles associated to the sequences ($\gamma_i^s$,$\beta_j^s$), red: individual frames}
\label{posebins}
\end{minipage}\hfill
\begin{minipage}{0.49\textwidth}
\centering
\includegraphics[width=\linewidth]{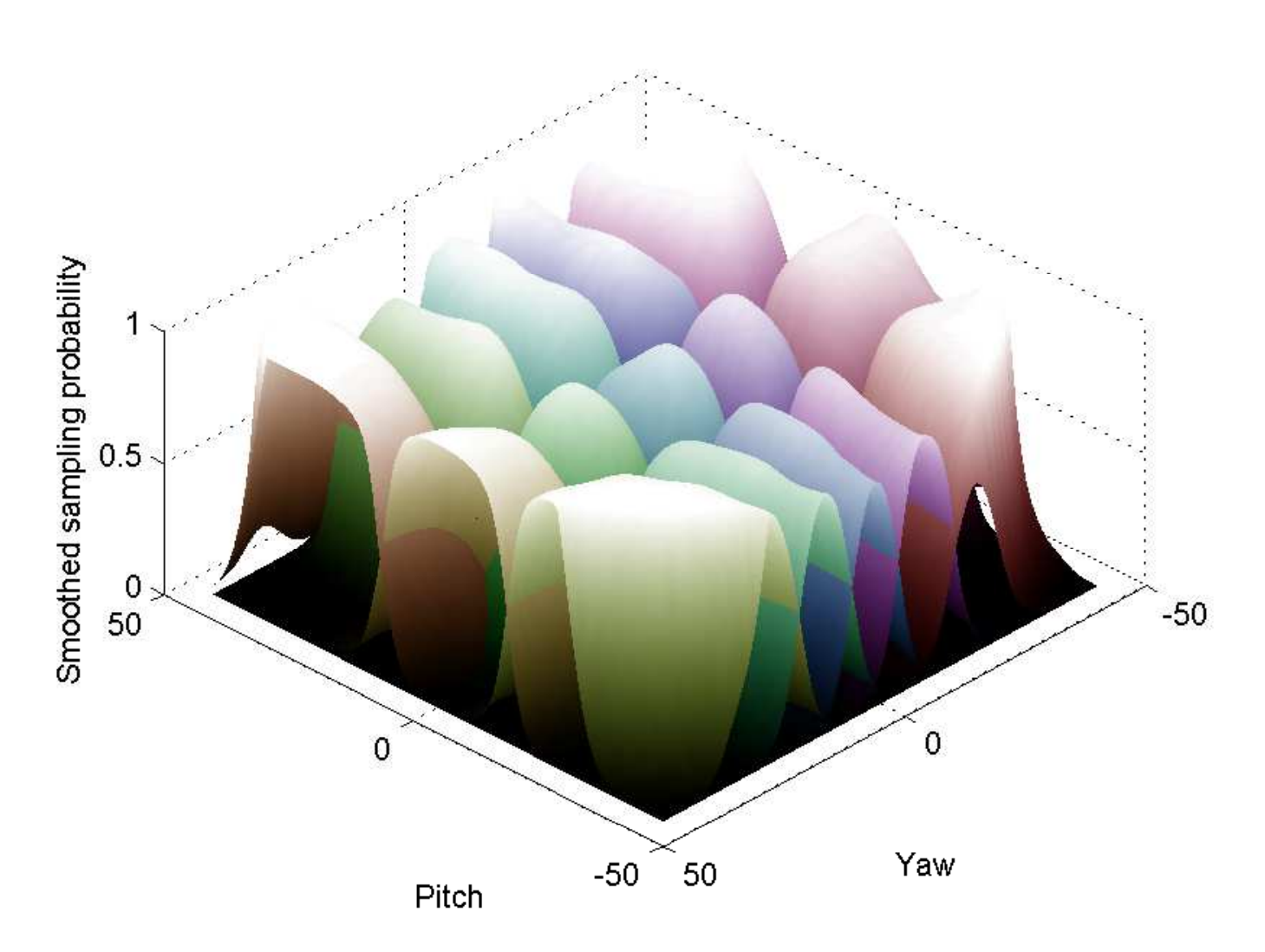}
\caption{Pose sampling probability distributions $\mathcal{P}_{\Omega_i}(\omega^n)$ constructed by smoothing the data repartition for each pose bin}
\label{smootheddistr}
\end{minipage}\hfill
\end{figure*}

Each texture frame of the BU-4DFE database is associated with a high-resolution $3D$ VRML model containing approximately $35000$ vertices, that we use to train our MVPCRF classifier as well as to design a new dataset for multi-view video FER. Many approaches \cite{tariq2012multi,vieriufacial} present results for static multi-view FER using the BU-3DFE database \cite{yin20063d}. To do that, for each static image, the authors typically render $3D$ meshes from a viewpoint with fixed yaw and pitch rotation angles. However, for video FER, head pose does not necessarily remain constant throughout a video. Furthermore, from the perspective of a fully automatic multi-view FER system, we typically aim at covering a specific head pose range rather than a discrete, arbitrary set of viewpoints. Hence, we propose to generate rotated versions of the videos by assigning each sequence a yaw-pitch variation from the frontal video. More specifically, our goal is to cover the same ``useful'' range as in \cite{tariq2012multi,vieriufacial} (\textit{i.e} $\pm 45$ yaw, $\pm 30$ pitch). We thus generate $k=5 \times 3$ bins $\{\Omega_i = \Omega_{\gamma_i,\beta_i}\}_{i = 1,...,k}$ with $\{\gamma_i\} = \{0,\pm17.5,\pm35\}$ and $\{\beta_i\} = \{0,\pm 25\}$ the mean rotation angles respectively in yaw and pitch. Each sequence $s$ is thus associated with rotation angles:

\begin{equation}\label{angle}
\begin{cases}
\ \gamma_i^s = \gamma_i + \gamma' \\
\ \beta_j^s = \beta_j + \beta'
\end{cases}
\end{equation}

Where $\gamma'$ and $\beta'$ are random variations uniformly drawn from the ranges $[-\sigma_{\gamma}, \sigma_{\gamma}]$ and $[-\sigma_{\beta}, \sigma_{\beta}]$, respectively. $\sigma_{\gamma}$ and $\sigma_{\beta}$ respectively denote the expected yaw and pitch width of the pose bins. In order to set those values, we measure the standard deviation of the head pose angles on the frontal view ($3.6$ and $5.9$ in yaw and pitch respectively). We then set $\sigma_{\gamma}=\sigma_{\beta}=5$ to allow a small overlap, thus a smoother interpolation between adjacent pose bins. The data distribution among the generated pose bins can be seen in Figure \ref{posebins}. For each frame of each sequence $s$, we generate $15$ frames by rotating the camera (position, direction and up vector). We also turn off the camera headlight and add an ambient light node to the VRML virtual environment.

The next step is to align facial feature points on the rotated sequences. However, the standard pipeline of applying a frontal or full profile face detection before aligning the feature points from the output face rectangle is bound to fail when the yaw/pitch becomes important and only a few images can correctly be aligned. In order to circumvent those issues, we generate ``boot" sequences using the first image of each video. Those sequences contain $20$ frames and show a very progressive rotation of the first frame starting from a frontal view and ending on the expected viewpoint. We apply the OpenCV Viola-Jones face detector \cite{viola2001rapid} on the first frame of the boot sequence (frontal view). Then we align facial feature points with the SDM tracker \cite{xiong2013supervised} on the retrieved face rectangle. Feature points are then tracked throughout the boot sequence. Once the \textit{boot} is completed, feature points are tracked on all the frames of the rotated expression videos (Figure \ref{boot}). Finally, we crop the facial images to a constant size based on the feature point location and generate a total of $906030$ images.

Lastly, we construct our multi-view training set by manually selecting the neutral and apical frames using the same subsets as in the frontal case. Also, in order to filter out the incorrectly aligned frames, we automatically discard the frames for which more than $5$ feature points do not lie on the facial mesh. Our final training set thus consists of $122623$ face images. Note however that we did not apply any manual check to remove the misaligned frames, or the ones for which the $3D$ models contain some distortions. The image generation process took about $5$ days to complete on an \textit{I7-4770} CPU on a \texttt{Matlab} environment. In order to ensure reproducibility of the results as well as to facilitate further research, we plan on releasing the peak frame annotation used for training the classifiers as well as the code for rendering the files and generating the boot and expression videos. For each of the retrieved frames, we use the posit algorithm \cite{dementhon1995model} to estimate head pose from the feature points. Such setting allows to use the same head pose estimation for training and testing, as compared to, e.g. constructing the pose sampling distribution from the ground truth generated positions. Then, we compute the pose sampling probability distribution for each pose bin $\mathcal{P}_{\Omega_i}(\omega(x^n))$ by applying a Gaussian smoothing on the training data repartition in the yaw/pitch space (Figure \ref{smootheddistr}). Thanks to the booting procedure discussed above, the number of training samples between the different pose bins is roughly equivalent. However, this might not be the case for other datasets, where constructing a sampling probability from the data offers the advantage to implicitly downweight the sampling of pose-specific trees relatively to the amount of training data.

\section{Experiments}\label{Experiments}

In this section, we report accuracies obtained on two different FER scenarios. In Section \ref{expmonov} we report comparisons between different classification models on two well-known frontal FER databases, the Extended Cohn-Kanade and BU-4DFE databases. Furthermore, in order to evaluate the capabilities of the learned models to generalize on spontaneous FER scenarios, we report classification results for cross-database evaluation on two spontaneous databases, namely the FG-NET FEED and BP4D databases. We highlight that our conditional formulation of dynamic integration substantially increases the recognition accuracy on such difficult tasks. Furthermore, in Section \ref{expMVPCRF} we also evaluate our approach on multi-view video FER scenarios. Finally, in Section \ref{RealTime} we show the real-time capacitbility of our system.

\subsection{Databases}\label{databases}

\textbf{The CK+} or \textbf{Extended Cohn-Kanade database \cite{lucey2010extended}} contains 123 subjects, each one associated with various numbers of expression records. Those records display a very gradual evolution from a \textit{neutral} class towards one of the 6 universal facial expressions described by Ekman \cite{ekman1971constants} (\textit{anger}, \textit{happiness}, \textit{sadness}, \textit{fear}, \textit{digust} and \textit{surprise}) plus the non-basic expression \textit{contempt}. Expressions are acted with no head pose variation and their duration is about 20 frames. From this dataset we use 309 sequences, each one corresponding to one of the six basic expressions, and use the three first and last frames from these sequences for training. We did not include sequences labelled as \textit{contempt} because CK+ contains too few subjects performing \textit{contempt} and other expressions to train the pairwise classifiers.

\textbf{The BU-4DFE database \cite{yin2008high}} contains 101 subjects, each one displaying 6 acted facial expressions with moderate head pose variations. Expressions are still prototypical but they are generally exhibited with much lower intensity and greater variability than in CK+, hence the lower baseline accuracy. Sequence duration is about 100 frames. As the database does not contain frame-wise expression annotations, we manually selected neutral and apex of expression frames for training. Each frame is associated with high-resolution $3D$ model data recorded using a Di3D device, that we use in our experiments to generate expression videos from multiple viewpoints.

\textbf{The BP4D database \cite{zhang2014bp4d}} contains 41 subjects. Each subject was asked to perform 8 tasks, each one supposed to give rise to one of the 8 spontaneous facial expressions (\textit{anger}, \textit{happiness}, \textit{sadness}, \textit{fear}, \textit{digust}, \textit{surprise}, \textit{embarrassment} or \textit{pain}). In \cite{zhang2014bp4d} the authors extracted subsequences of about 20 seconds for manual FACS annotations, arguing that these subsets contain the most expressive behaviors.

\textbf{The FG-NET FEED database \cite{wallhoff2006database}} contains 19 subjects, each one recorded three times while performing 7 spontaneous expressions (the six universal expressions, plus the \textit{neutral} one). The data contain low-intensity emotions, very short expression displays, as well as moderate head pose variations.

\subsection{Experimental setup}\label{expsetup}

\subsubsection{Evaluation framework}\label{evalfw}

7-class RF (static) and PCRF (full and conditional) models are trained on the CK+ and BU-4DFE datasets using the set of hyperparameters described in Table \ref{hyperparameters}. Note however that extensive testing showed that the values of these hyperparameters had a very subtle influence on the performances. This is due to the complexity of the RF framework, in which individually weak trees (e.g. that are grown by only examining a few features per node) are generally less correlated, still outputting decent predictions when combined altogether. Nevertheless, we report those settings for reproducibility concerns. Also, for a fair comparison between static and pairwise models, we use the same total number of feature evaluations for generating the split nodes. For every test, we report results averaged over $5$ different runs, with a standard deviation lower than $0.25\%$ between each run.

\begin{table}[!ht]
\centering
\caption{Hyperparameter settings}
\begin{tabular}{| l | c | c |}
  \hline
	\rowcolor{Gray0}
	Hyperparameters & value(RF) & value(PCRF) \\
	\hline
	Nb. of $\phi^{(1)}$ features & 40 & 20 \\
	\hline
	Nb. of $\phi^{(2)}$ features & 40 & 20 \\
	\hline
	Nb. of $\phi^{(3)}$ features & 160 & 80 \\
	\hline
	Nb. of $\phi^{(4)}$ features & - & 20 \\
	\hline
	Nb. of $\phi^{(5)}$ features & - & 20 \\
	\hline
	Nb. of $\phi^{(6)}$ features & - & 80 \\
	\hline
	Data ratio per tree & $2/3$ & $2/3$ \\
	\hline
	Nb. of thresholds & 25 & 25 \\
	\hline
	Total nb. of features & 6000 & 6000 \\
	\hline
	Nb. of trees & 500 & 500 \\
	\hline
\end{tabular}
\label{hyperparameters}
\end{table}

During the evaluation, the prediction is initialized in a fully automatic way from the first frame using the static classifier. Then, for the full and conditional models, probabilities are estimated for each frame using transitions from previous frames only, bringing us closer to a real-time scenario. However, although it uses transitional features, our system is essentially a frame-based classifier that outputs an expression probability for each separate video frame. This is different from, for example, a HMM, that aims at predicting a probability related to all the video frames. Thus, in order to evaluate our classifier on video FER tasks, we acknowledge correct classification if the maximum probability outputted for all frames corresponds to the ground truth label. This evaluates the capability of our system to retrieve the most important expression mode in a video, as well as the match between the retrieved mode and the ground truth label.

For the tests on CK+ and BU-4DFE databases, both static and transition classifiers are evaluated using the Out-Of-Bag (OOB) error estimate \cite{breiman2001random}. More specifically, bootstraps for individual trees of both static and pairwise classifiers are generated at the subject level. Thus, during evaluation, each tree is applied only on subjects that were not used for its training. The OOB error estimate is an unbiased estimate of the true generalization error \cite{breiman2001random} which is faster to compute than Leave-One-Subject-Out or $k$-fold cross-evaluation estimates. Also, it has been shown to be generally more pessimistic than traditional error estimates \cite{bylander2002estimating}, further empathizing the quality of the proposed contributions.

\subsection{FER from frontal view videos}\label{expmonov}

\begin{figure}[ht]
\centering
\includegraphics[width=\linewidth]{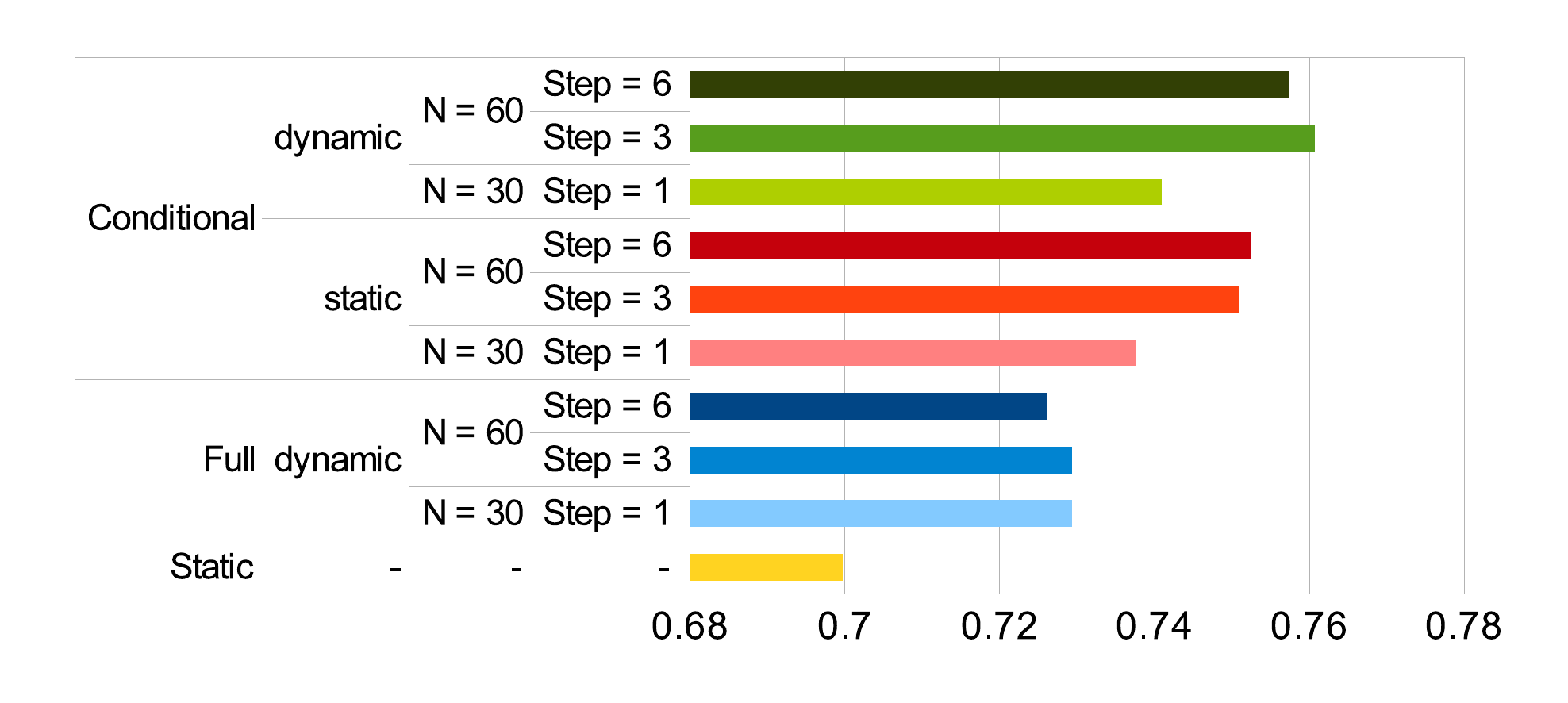}
\caption{Average accuracy rates obtained for various temporal integration parameters on the BU-4DFE database}
\label{compBU4DFE}
\end{figure}

\begin{figure*}[ht]
\centering
\includegraphics[width=\linewidth]{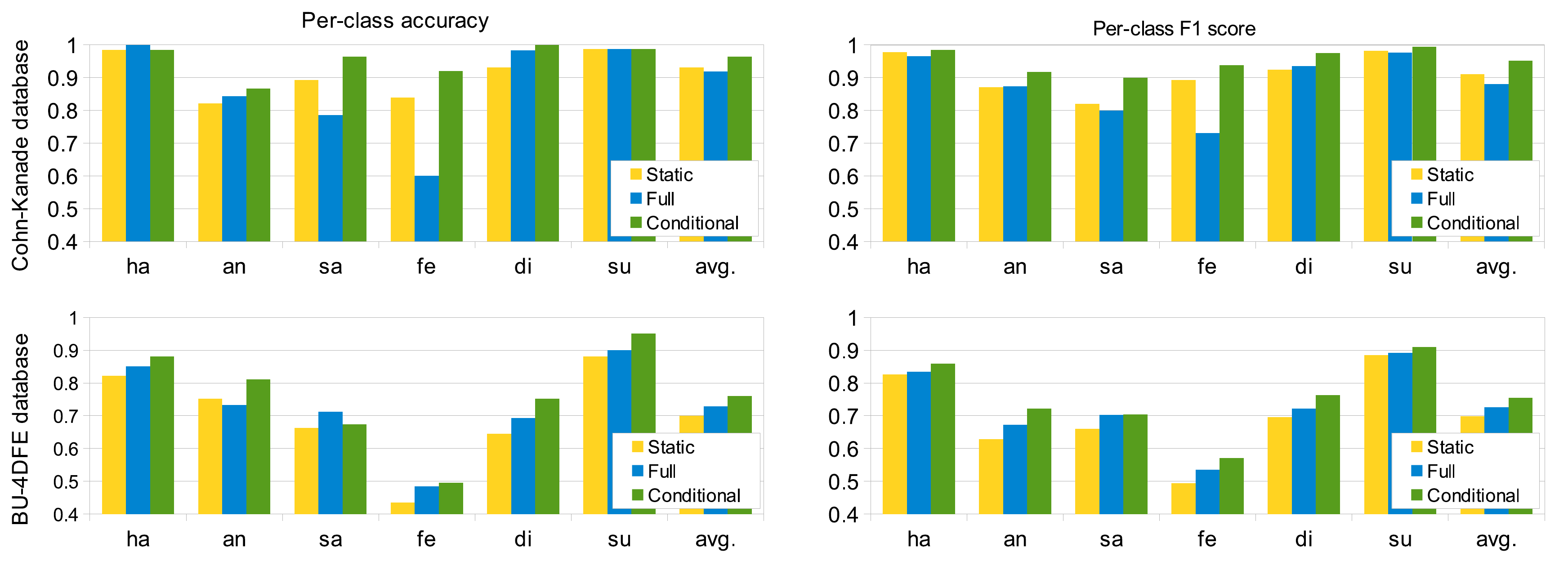}
\caption{Per-class recognition accuracy rates and F1-scores on CK+ and BU-4DFE databases}
\label{perclassPCRF}
\end{figure*}

In order to validate our approach on frontal view videos, we compared our conditional model to a purely static model and a full model, for a variety of dynamic integration parameters (the length of the temporal window $N$ and the step between those frames \textit{Step}) on the BU-4DFE database. We also evaluated the interest of using a \textit{dynamic} probability prediction for previous frames (\textit{i.e.} the output of the pairwise classifier for those frames) versus a \textit{static} one. Average results are provided in Figure \ref{compBU4DFE}. For CK+ database, sequences are generally too short to show significant differences when varying the temporal window size or the step size. Thus we only report accuracy for full and conditional models with a window size of 30 and a step of 1. Per-expression accuracies and F1-scores for both Cohn-Kanade and BU-4DFE databases are shown in Figure \ref{perclassPCRF}.

Figure \ref{perclassPCRF} reveals that facial expressions involving large deformations (e.g. $surprise$ and $happy$) are recognized with very high accuracies. $Disgust$ is also recognized quite well for both databases and for all the models. However, more subtle expressions such as $anger$ and $sadness$ rank among the lowest. For those expressions, the addition of spatio-temporal allows to increase the recognition accuracy as compared to a static RF model. As in many other works on facial expressions, accuracies for $fear$ are lower than for the other expressions, as it can be quite subtle in some cases where the eyes are open a little bit wider. Moreover, this expression also displays larger variability than the others on these databases. Overall, modelling transition patterns through PCRF allows to significantly increase the recognition accuracy as well as the balanced $F1$-score, for all expressions on both CK+ and BU-4DFE databases. We believe that this is due to the extra dynamic features that provide both robustness and decorrelation of the individual decision trees.

Figure \ref{perclassPCRF} also shows that the conditional model outperforms the full model on both databases, which is probably due to the fact that using only a restricted set of ongoing expression transitions for training allows to better capture the variability of the spatio-temporal features for the dedicated pairwise forests. This is particularly true on the CK+ database, where the number of pairwise data points is not enough for the full model to capture the variability of all possible ongoing transitions, hence justifying the lower accuracy. Table \ref{compBU4DFE} also shows that it is better to look backward for more frames in the sequence ($N$ = 60) with less correlation between the frames (\textit{Step} = 3 or 6). Again, such setting allows to take more decorrelated paths in the individual trees, giving a better recombination after averaging over time.

A compilation of comparisons to other state-of-the-art approaches for FER can be found in Table \ref{compsota}. On the CK+ dataset, we compare our algorithms with recent works reporting results on the same subset of sequences (\textit{i.e.} not including \textit{contempt}). Such comparisons are to be put into perspective as the evaluation protocols differ between the methods. Nevertheless, PCRF provides slightly better results than those reported in \cite{mohammadi2014non} ($+3.2\%$) as well as in \cite{shojaeilangari2014multi} ($+1.9\%$) and \cite{happyautomatic} ($+2.3\%$). Furthermore, those approaches explicitly perform normalization w.r.t. a neutral face and consider the last (apex) frame whereas our approach automatically retrieves the apex as the maximum probability throughout a sequence. 

Moreover, to the best of our knowledge, our approach gives the best results on the BU-4DFE database for automatic FER from videos using $2D$ information only. It provides better results than the dynamic $2D$ approach \cite{sun2008facial} ($+9.1\%$), as well as the LBP-TOP approach presented in \cite{hayat2012evaluation} ($+4.5\%$). Recently, Meguid \textit{et al.} \cite{abd2014fully} obtained satisfying results using an original hybrid RF/SVM system. They trained on the static BU-3DFE database \cite{yin20063d} and employ a post-classification temporal integration scheme. However our PCRF method achieved a significantly higher accuracy ($+3\%$) which shows the benefits of using dynamic information at the feature level. 

\begin{table}[!ht]
\centering
\caption{Comparisons with state-of-the-art approaches}
\begin{tabular}{| l | r |}
	\hline
	\rowcolor{Gray0}
	CK+ database & Accuracy\\
	\hline
	Mohammadi \textit{et al.} \cite{mohammadi2014non} & 93.2\\
	\hline
	Happy \textit{et al.} \cite{happyautomatic} & 94.1\\
	\hline
	Shojaeilangari \textit{et al.} \cite{shojaeilangari2014multi} & 94.5 \\
	\hline
	This work, RF & 93.2 \\
	\hline
	This work, PCRF & \textbf{96.4} \\
	\hline
	\rowcolor{Gray0}
	BU-4DFE database & Accuracy\\
	\hline
	Sun \textit{et al.} \cite{sun2008facial} & 67.0 \\
	\hline
	Hayat \textit{et al.} \cite{hayat2012evaluation} & 71.6 \\
	\hline
	Meguid \textit{et al.} \cite{abd2014fully} & 73.1 \\
	\hline
	This work, RF & 70.0 \\
	\hline
	This work, PCRF & \textbf{76.1} \\
	\hline
	\rowcolor{Gray0}
	FEED database (Cross-db) & Accuracy\\
	\hline
	Meguid \textit{et al.} \cite{abd2014fully} & 53.7 \\
	\hline
	This work, RF & 51.9 \\
	\hline
	This work, PCRF & \textbf{57.1}\\
	\hline
	\rowcolor{Gray0}
	BP4D database (Cross-db) & Accuracy\\
	\hline
	Zhang \textit{et al.} \cite{zhang2014bp4d} & 71.0 \\
	\hline
	This work, RF & 68.6 \\
	\hline
	This work, PCRF & \textbf{76.8} \\
	\hline
\end{tabular}
\label{compsota}
\end{table}

Table \ref{compsota} also reports results for cross-database evaluation (with training on the BU-4DFE database) on the FEED database. In order to provide a fair comparison between our approach and the one presented in \cite{abd2014fully}, we used the same labelling protocol. One can see that the performances of their system are better than those of our static RF model, which can be attributed to the fact that they use a more complex classification and posterior temporal integration flowchart. Nevertheless, our PCRF model provides a substantially higher accuracy ($+3.4\%$), which, again, is likely to be due to the use of spatio-temporal features as well as an efficient conditional integration scheme. Furthermore, modelling spatio-temporal patterns for \textit{every} possible transition (\textit{i.e.} across the videos) allows to gather more training data than using spatio-temporal descriptors \cite{zhao2007dynamic, klaser2008spatio} learnt on separate videos.

We also performed cross-database evaluation on the BP4D database. Again, for a fair comparison, we used the same protocol as in \cite{zhang2014bp4d}, with training on the BU-4DFE database and using only a subset of the tasks (\textit{i.e.} tasks 1 and 8 corresponding to expression labels \textit{happy} and \textit{disgust} respectively). However, we do not retrain a classifier with a subset of 3 expressions as it is done in \cite{zhang2014bp4d}, but instead use our 7-class static and PCRF models with a forced choice between \textit{happiness} (probability of class \textit{happiness}) and \textit{disgust} (probability sum of classes \textit{anger} and \textit{disgust}). Such setting could theoretically increase the confusion in our conditional model, resulting in a lower accuracy. However, as can be seen in Table \ref{compsota}, using dynamic information within the PCRF framework allows to substantially increase the recognition rate as compared to a static RF framework ($+8.2\%$). We also overcome the results reported in \cite{zhang2014bp4d} by a significant margin ($+5.8\%$), further showing the capability of our approach to deal with complex spontaneous FER tasks. Also note that in \cite{zhang2014bp4d}, the authors used the so-called \textit{Nebulae} $3D$ polynomial volume features which are by far more computationally expensive than our geometric and integral HOG $2D$ features. All in all, we believe our results show that the PCRF approach provides significant improvements over a traditional static classification pipeline that translates very well to more complicated spontaneous FER scenarios, where a single video may contain samples of several expressions.

\subsection{Multi-view experiments}\label{expMVPCRF}

To the best of our knowledge, there is currently no publicly available benchmark for evaluating dynamic FER methods under yaw/pitch head pose variations. We thus propose a new evaluation protocol using the rotated videos generated in Section \ref{datapreparation}. During evaluation, for each frame $n$ of a sequence, head pose $\omega(x^n)$ is thus estimated from the set of aligned feature points and trees from the MVPCRF collections are sampled according to the values $\mathcal{P}_{\Omega_i}(\omega(x^n))$ for each pose bin $\Omega_i$. We compare the average accuracies outputted by RF, PCRF, MVRF and MVPCRF. RF and PCRF were trained on the central (frontal view) bin only. For PCRF and MVPCRF, we set the temporal integration parameters $N=60$ and $Step=6$ as it provided satisfying results in the frontal case (Figure \ref{compBU4DFE}). As in Section \ref{expmonov}, a video is considered correctly classified if the dominant expression mode (\textit{i.e.} the maximum probability expression throughout the sequence) corresponds to the ground truth label for that video.

Table \ref{perclassaccmv} displays per-expression accuracies averaged over the $15$ pose bins for the three models. For all expressions, MVPCRF outperforms RF and PCRF by a significant margin. MVPCRF also outperforms the static multi-view MVRF on all expressions but \textit{sadness} and \textit{fear}. However, Table \ref{perclassf1mv} reveals that the F1-score is a little higher for MVPCRF on those expressions, indicating that the static MVRF is more biased toward those expression classes. This seems particularly relevant in the positive pitch case, where using spatio-temporal information helps to disambiguate \textit{anger} from \textit{sadness}, which in some case differ only by a very subtle eyebrow frown or lip raiser. Also, \textit{fear} appears as the most subtle expression as already reported in other works \cite{abd2014fully}. This is due to the fact that subjects often smile during the sequence, thus the videos may be misclassified as \textit{happiness}. For this reason, many other approaches such as the one in \cite{ben20144} use a restricted number of subjects. However, we use the $101$ subjects to ensure reproducibility of the results.

The overall classification accuracy is $72.2\%$ against $76.1\%$ for the benchmarks of Section \ref{expmonov} on frontal view video. This performance drop comes from a greater variability in face appearance as well as the feature point misalignment for non-frontal poses, as discussed in \cite{jeni2011high}. Classification rates are also a little lower than the static FER baseline \cite{tariq2012multi,vieriufacial} on the BU-3DFE database. However, fully automatic FER from video is a much more difficult setup, as it involves the retrieval of the apex frames and expression classification on those frames. Furthermore, many approaches operate on high-resolution $3D$ data and require expensive projections on a frontal view, thus can not be applied easily to real-time FER from consumer camera.

\begin{figure*}[ht]
\centering
\includegraphics[trim=3cm 0 3cm 0,width=\linewidth]{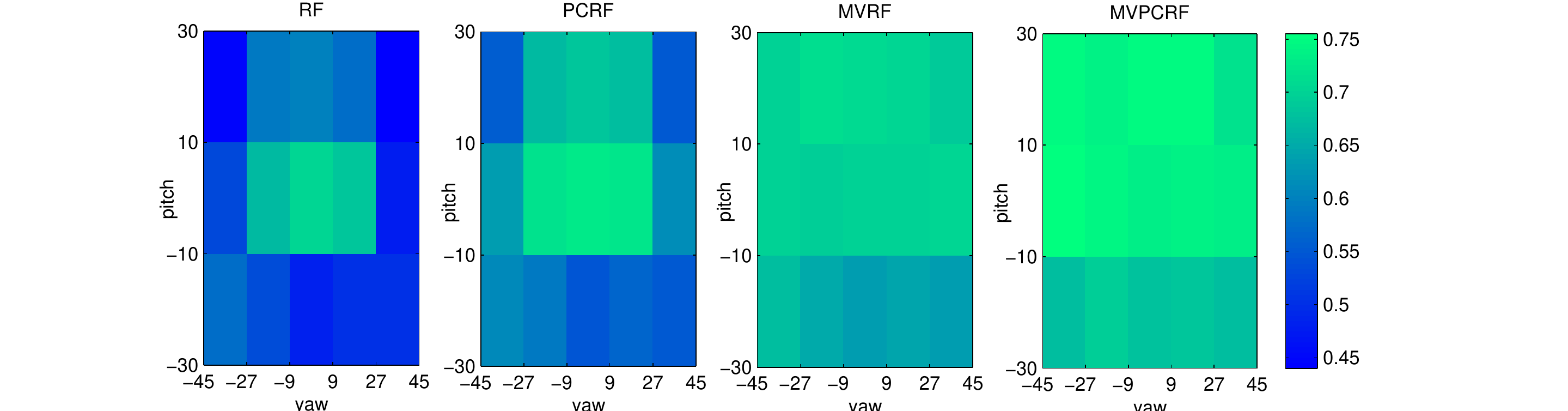}
\caption{Per-pose bin accuracy rates averaged over all expressions}
\label{avgMv}
\end{figure*}

Figure \ref{avgMv} shows the per-pose bin accuracy rates averaged over the six expressions. On the one's hand, RF performances seems to drop dramatically when we move away from the central bin (from $70.4\%$ to $44.7\%$). Interestingly, PCRF performs significantly better than RF on every pose bin, which proves that the captured dynamics generalize well on unseen data, as already shown on the cross-database settings. PCRF performance also drops significantly on off-center pose bins. On the other hand, MVPCRF performs significantly better on those bins: accuracy is nearly symetrical for negative and positive yaws, as already reported by \cite{tariq2012multi} for static multi-view FER. Furthermore, as stated in \cite{tariq2012multi,vieriufacial} we observe lower classification rates on negative pitches ($68.3\%$ as compared to $74.1\%$ on average for positive pitch). Our take is that the mouth area may be the most informative one for FER tasks: as such, the classifiers can struggle to disambiguate certain expressions (e.g. \textit{anger} from \textit{sadness}) when the mouth features become more subtle and difficult to capture.  

\begin{table*}
\centering
\begin{minipage}{0.5\textwidth}
\caption{Per-expression accuracy rates averaged over all pose bins}
\label{perclassaccmv}
\scalebox{0.65}{
\begin{tabular}{| c | c | c | c | c |}
  \hline
	\rowcolor{Gray0}
	Expression & RF (\%) & PCRF (\%) & MVRF (\%) & MVPCRF (\%)\\
	\hline
	\cellcolor{Gray0} Happy & \cellcolor{Blue3} 57.8 & \cellcolor{Blue23} 73.4 & \cellcolor{Blue2} 83.3 & \cellcolor{Blue1}\textbf{87.8}\\
	\cellcolor{Gray0} Angry & \cellcolor{Blue3} 59.2 & \cellcolor{Blue2} 73.3 & \cellcolor{Blue2} 71.9 & \cellcolor{Blue12} \textbf{80.4}\\
	\cellcolor{Gray0} Sad & \cellcolor{Blue34} 56.0 & \cellcolor{Blue34} 52.2 & \cellcolor{Blue23} \textbf{70.8} & \cellcolor{Blue23} 64.4\\
	\cellcolor{Gray0} Fear & \cellcolor{Blue45} 29.6 & \cellcolor{Blue45} 25.7 & \cellcolor{Blue4} \textbf{34.8} & \cellcolor{Blue4} 33.0\\
	\cellcolor{Gray0} Disgust & \cellcolor{Blue34} 48.4 & \cellcolor{Blue3} 63.9 & \cellcolor{Blue3} 63.5 & \cellcolor{Blue2} \textbf{74.3}\\
	\cellcolor{Gray0} Surprise & \cellcolor{Blue2} 81.6 & \cellcolor{Blue1} 88.3 & \cellcolor{Blue12} 85.3 & \cellcolor{Blue1} \textbf{92.4}\\
	\hline
	\cellcolor{Gray0} Average & \cellcolor{Blue34} 55.4 & \cellcolor{Blue3} 62.8 & \cellcolor{Blue23} 68.3 & \cellcolor{Blue2}\textbf{72.1}\\
	\hline
\end{tabular}}
\end{minipage}
\begin{minipage}{0.49\textwidth}
\caption{Per-expression F1-scores averaged over all pose bins}
\label{perclassf1mv}
\scalebox{0.65}{
\begin{tabular}{| c | c | c | c | c |}
  \hline
	\rowcolor{Gray0}
	Expression & RF (\%) & PCRF (\%) & MVRF (\%) & MVPCRF (\%)\\
	\hline
	\cellcolor{Gray0} Happy & \cellcolor{Blue3} 62.6 & \cellcolor{Blue2} 74.4 & \cellcolor{Blue12} 80.7 & \cellcolor{Blue1} \textbf{84.2}\\
	\cellcolor{Gray0} Angry & \cellcolor{Blue4} 48.6 & \cellcolor{Blue3} 61.5 & \cellcolor{Blue3} 62.9 & \cellcolor{Blue23} \textbf{68.0}\\
	\cellcolor{Gray0} Sad & \cellcolor{Blue4} 46.2 & \cellcolor{Blue4} 48.8 & \cellcolor{Blue23} 65.4 & \cellcolor{Blue23} \textbf{66.1}\\
	\cellcolor{Gray0} Fear & \cellcolor{Blue45} 34.7 & \cellcolor{Blue45} 34.7 & \cellcolor{Blue34} 43.8 & \cellcolor{Blue34} \textbf{44.8}\\
	\cellcolor{Gray0} Disgust & \cellcolor{Blue3} 56.3 & \cellcolor{Blue23} 66.2 & \cellcolor{Blue23} 67.9 & \cellcolor{Blue2} \textbf{73.1}\\
	\cellcolor{Gray0} Surprise & \cellcolor{Blue23} 71.4 & \cellcolor{Blue2} 77.6 & \cellcolor{Blue12} 83.6 & \cellcolor{Blue1} \textbf{87.3}\\
	\hline
	\cellcolor{Gray0} Average & \cellcolor{Blue34} 53.3 & \cellcolor{Blue3} 60.6 & \cellcolor{Blue23} 67.4 & \cellcolor{Blue2} \textbf{70.6}\\
	\hline
\end{tabular}}
\end{minipage}
\end{table*}

\begin{figure*}[ht]
\centering
\includegraphics[trim=4cm 0 0 0,width=1.2\linewidth]{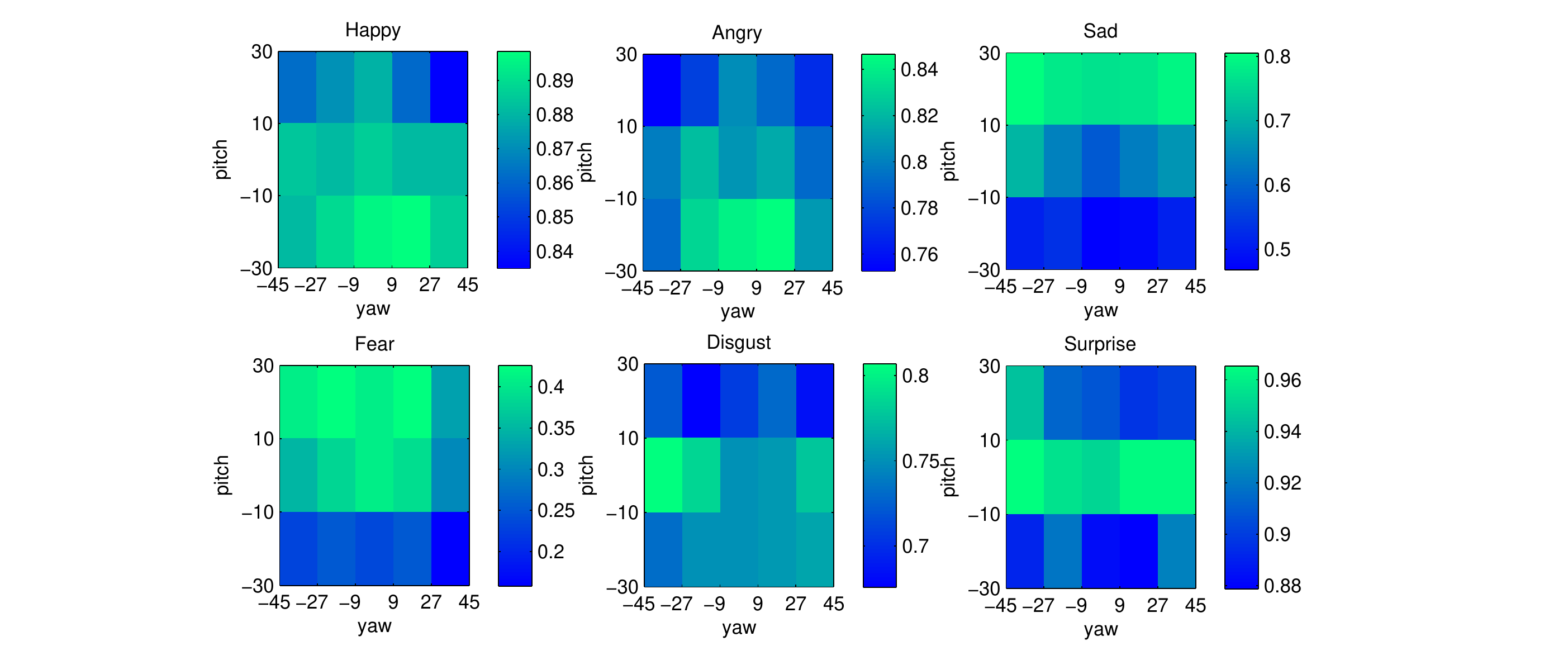}
\caption{Per-expression, per-pose bin classification accuracy rates}
\label{perclassMV}
\end{figure*}

Figure \ref{perclassMV} shows per-expression, per-pose bin accuracies obtained for MVPCRF. Indeed, expressions such as \textit{sadness} and \textit{fear} are better recognized for positive pitches, as they specifically involve subtle mouth movements as well as eyebrow raising. Conversely, \textit{anger} and \textit{disgust} are characterized by eyebrow frowning that is better recognized on negative pitch views. Finally, \textit{happiness} and \textit{surprise} are expressions with the highest overall classification rates. They are typically better recognized on frontal views or for negative pitches, where the corresponding mouth motions are less frequently misclassified as \textit{fear}.

\subsection{Complexity analysis}\label{RealTime}

An advantage of using conditional models is that with equivalent parallelization they are faster to train than a full model learnt on the whole dataset. According to \cite{louppe2014understanding} the average complexity of training a RF classifier with $M$ trees is $\mathcal{O}(MKN\log^2{}N)$, with $K$ being the number of features to examine for each node and $N$ the size of (2/3 of) the dataset. Thus if the dataset is equally divided into $P$ bins of size $\tilde{N}$ upon which conditional forests are trained (and such that $N = P\tilde{N}$), the average complexity of learning a conditional model now becomes $\mathcal{O}(MKN\log^2{}\tilde{N})$.

Same considerations can be made concerning the evaluation, as trees from the full model are bound to be deeper than those from the conditional models. Table \ref{time} shows an example of profiling a MVPCRF on one video frame with an averaging over 60 frames and a step of 6 frames. We experiment with various total numbers of trees $M$ to show that the proposed framework can perform real-time FER. 

\begin{table}[!ht]
\centering
\caption{Profiling of total processing time for one frame (in ms)}
\begin{tabular}{| l | c |}
  \hline
	\rowcolor{Gray0}
	Step & Time (ms) \\
	\hline
	Facial alignment & 10.0\\
	\hline
	Integral HOG channels computation & 2.0\\
	\hline
	MVPCRF evaluation ($M$ = 500) & 2.6\\
	\hline
	MVPCRF evaluation ($M$ = 1000) & 4.8\\
	\hline
	MVPCRF evaluation ($M$ = 2000) & 7.8\\
	\hline
	MVPCRF evaluation ($M$ = 6000) & 19.0\\
	\hline
\end{tabular}
\label{time}
\end{table}

This benchmark was conducted on a \textit{I7-4770} CPU within a C++/OpenCV environment, without any code parallelization. As such, the algorithm already runs in real-time. Furthermore, evaluations of pairwise classification or tree subsets can be parallelized to fit real-time processing requirements on low-power engines such as mobile phones. In addition, the facial alignment step can be performed at more than 300 fps on a smartphone with similar performances using the algorithms from \cite{ren2014face}.

\section*{Discussion and Conclusion}\label{ccl}

In this paper, we presented an adaptation of the Random Forest framework for automatic dynamic pose-robust facial expression recognition from videos. We also introduced a novel way of integrating the temporality of expressions by considering pairwise RF classifiers. This formulation allows the efficient integration of high-dimensional, low-level spatio-temporal information through averaging over time pairwise trees. These trees are conditioned on predictions outputted for the previous frames to help reducing the variability of the ongoing transition patterns. In addition, we proposed an extension of the PCRF framework to efficiently handle head pose variation in an expression recognition system. We showed that our models can be trained and evaluated efficiently given appropriate data, and lead to a significant increase of performances compared to a static RF. We also introduced a new multi-view video corpus generated using the BU-4DFE database to assess the pose-robustness of the proposed system. The \texttt{Matlab} code used to render the images that we used for training and testing the classifiers will be made publicly available. Finally, we showed that our method works on real-time without specific optimization schemes, and could be run on low-power architectures such as mobile phones by using appropriate paralellization scheme.

Nevertheless, the proposed algorithms are still not perfect and suffer from a number of limitations. First, In order to train a PCRF, one need to explicitly consider frame-level annotations. In our work, we manually highlighted a set of peak frames for the database, which were used to train the classifiers. This is a recurrent drawback of frame-based classifiers as compared to sequence-level ones (e.g. HMMs, CRFs) and thus could not be solved easily. However, using only a subset of the videos for training also allows to limit the memory usage, which is particularly relevant in our case, as we need to store the feature maps for each image. Moreover, we demonstrated that, during evaluation, our algorithm was successful at retrieving the correct expression modes and thus could be applied in a fully automatic fashion. Moreover, an advantage of integrating temporality under the form of transition modelling is that it does not require continuity of the sequence. Hence, it has no problem handling failure from the detection or feature point alignment pipelines, as opposed to other spatio-temporal descriptors \cite{zhao2007dynamic,klaser2008spatio}. Secondly, in order to build transition classifiers we need examples for each possible transition. This can be a hindrance when training on highly unbalanced datasets (as in CK+ with \textit{contempt} expression class). Thirdly, multi-view classification requires loads of training data from multiple head poses. This, however, can be alleviated by the use of high-dimensional $3D$ models to generate training examples. Finally, we felt that, particularly for the experiments on multi-view data, we were at times limited by the robustness of the feature point alignment for non-frontal head pose, as well as from the quality of the pose estimation from the set of feature points. Such problem could in theory be alleviated by the use of recent robust algorithms such as the one proposed in \cite{xiong2015global}.

As such, future work will consist in addressing  self-occlusions to adapt the proposed MVPCRF framework to other spontaneous or in the wild FER datasets such as the AFEW database \cite{dhall2011acted}. As $3D$ models have already been very useful for generating data for a variety of problems, we believe that robustness to self-occlusions could also benefit from such data. Furthermore, we would like to investigate applications of transition modelling for other video classification/regression problems such as Facial Action Unit intensity prediction or body and hand gesture recognition.

\bibliographystyle{unsrt}
\bibliography{Biblio}   




\end{document}